\documentclass[10pt,twocolumn,letterpaper]{article}

\usepackage{cvpr_arxiv}
\usepackage{times}
\usepackage{epsfig}
\usepackage{graphicx}
\usepackage{amsmath}
\usepackage{amssymb}

\usepackage{color}
\usepackage[pagebackref=true,breaklinks=true,letterpaper=true,colorlinks,bookmarks=false]{hyperref}

\cvprfinalcopy 

\def\cvprPaperID{****} 

\ifcvprfinal\fi
\begin{document}

\title{Privacy Leakage Avoidance with Switching Ensembles}

\author{%
  Rauf Izmailov \\
  Perspecta Labs\\
  Basking Ridge, NJ 07920\\
  {\tt\small rizmailov@perspectalabs.com}
\and
  Peter Lin \\
 Perspecta Labs\\
  Basking Ridge, NJ 07920\\
  {\tt\small plin@perspectalabs.com}
\and
  Chris Mesterharm \\
 Perspecta Labs\\
  Basking Ridge, NJ 07920\\
  {\tt\small jmesetrharm@perspectalabs.com}
\and
  Samyadeep Basu \\
 Department of Computer Science\\
 University of Maryland\\
 College Park, MD 20740\\
  {\tt\small sbasu12@cs.umd.edu}
}

\maketitle

\begin{abstract}
We consider membership inference attacks, one of the main privacy
issues in machine learning. These recently developed attacks have
been proven successful in determining, with confidence better than a
random guess, whether a given sample belongs to the dataset on which
the attacked machine learning model was trained. Several approaches
have been developed to mitigate this privacy leakage but the
tradeoff performance implications of these defensive mechanisms
(i.e., accuracy and utility of the defended machine learning model)
are not well studied yet. We propose a novel approach of privacy
leakage avoidance with switching ensembles (PASE), which both
protects against current membership inference attacks and does that
with very small accuracy penalty, while requiring acceptable
increase in training and inference time. We test our PASE method,
along with the the current state-of-the-art PATE approach, on three
calibration image datasets and analyze their tradeoffs.
\end{abstract}

\section{Introduction}

The spectacular successes of machine learning in recent years have
also brought significant scrutiny in its high-profile security
vulnerabilities. Although the focus has been mostly on adversarial
attacks, which were successfully constructed for a variety of
applications~\cite{Biggio13, Biggio14, Nelson12, Papernot16a,
  Papernot17, Moosavi16, Papernot16b, Kurakin16, Sharif16}, another
group of security issues was recently identified in machine learning
concerning various forms of privacy attacks. It was shown that it is
possible to extract information about the training data by analyzing
or querying a machine learning
model~\cite{Ateniese,Fredrikson1,Fredrikson2,Wu,Basu2019MembershipMI,Tramer}.

In this paper, we focus on a type of attack commonly referred to as
membership inference attack. A successful membership inference attack
can determine, with better than random success rate, if a given sample
belongs to the training set when given access to a machine learning
model~\cite{Shokri}. This type of attack was introduced
in~\cite{Shokri}, and later~\cite{Shokri2} proposed a method of
providing privacy guarantees while taking into account the accuracy of
the defended classifier. The attack~\cite{Shokri} leverages the
differences in the target model's predictions (confidence values) on
the samples it was trained on versus the samples which were outside of
training data for detecting if a sample was part of the training data.
Even with such limited information, this attack can be effective and
remains effective for a variety of privacy defense methods~\cite{Shokri}.

There have been many recent research advances in the area of
membership inference attacks.  While~\cite{Shokri} makes several
assumptions that weaken the scope of membership attacks such as
knowledge of the type of model and knowledge of certain aspects of the
training data,~\cite{ml_leaks} is able to generate successful attacks
with weaker assumptions.  In~\cite{member_explain}, membership attacks
exploit interpretable machine learning output to get extra information
about the training data.  Even unsupervised learning is susceptible as
\cite{generative} has described successful membership attacks in case
of generative adversarial networks.  Perhaps the most interesting
research has been applying differential privacy~\cite{dp} as a way to
guarantee a defense against membership attacks.  The PATE algorithm
\cite{dp_attack, pate} uses a teacher-student framework to protect any
machine learning algorithm from a broad range of privacy attacks,
including membership attacks, with differential privacy guarantees.

However, a security solution (in this context, privacy protection)
often entails some performance penalty (deterioration of accuracy
and / or utility of the defended system). For instance, for the case
of preventing privacy leakage of machine learning models, an obvious
approach is to restrict the classifier's output to discrete labels
without disclosing the associated confidence values. However, this
approach significantly reduces the utility of machine learning model
since the choice of the subsequent actions frequently depends on the
level of confidence of the provided classification decision. A less
obvious approach is to restrict the class of machine learning models
to simple classifiers such as linear threshold functions as their
output does not have local optima that can betray the identity of
training data. However, the applicability of such classifiers to
realistic, complex problems is severely limited. Finally, approaches
involving strong differential privacy guarantees can negatively
affect the utility / performance of the protected model. In the case
of PATE~\cite{dp}, the algorithm depends on an ensemble of teachers
where each teacher is based on a small subset of the original
training data.  This not only negatively affect the resulting
accuracy, but also does that in the circumstances where the problem
of privacy leakage is most serious. Specifically, our experiments
suggest that the efficiency of privacy attacks increases with the
decrease of the size of training dataset since that makes training
data more ``sparse'' in the feature space, which exposes more data
``individuality'' that can then be then detected more easily by the
attacker.

In this paper, we propose and investigate a novel framework of
protecting machine learning models against membership attacks. Our
framework is based on ensemble learning with the focus on
maintaining utility and accuracy of the existing machine learning
model that is being protected. For this framework, we modify the
generally solid approach of constructing an ensemble of classifiers
by replacing its final step of combining {\em all} their outputs
(using, for instance, some weighted voting) with a switching
decision that selects {\em only one} of the classifiers in the
ensemble for providing the classification output for a given input.
This modification allows us to train the classifiers of the ensemble
on {\em significantly larger} subsets of the original dataset, thus
incurring negligible performance penalty while maintaining the
desired privacy protection of the training data identity. We
describe the architecture of our proposed Privacy Leakage Avoidance
with Switching Ensembles (PASE) method, discuss its properties and
implementation details, and calibrate its performance (along with
the state-of-the-art PATE approach~\cite{pate}) on three popular
image datasets (CIFAR-10~\cite{CIFAR10}, MNIST~\cite{MNIST}, SVNH
~\cite{SVNH}) within the framework of the state-of-the-art
membership inference attack proposed in~\cite{Shokri}.

\section{Privacy Leakage Avoidance Method}

The {\em training phase} of PASE consists of building a set
(ensemble) of individual auxiliary classifiers, each trained on a
large subset of the original training data, where these subsets are
constructed in the same way as subsets for cross-validation; the
accuracy of each classifier would thus be close to that of the
baseline machine learning model that uses all the available data for
its training. The {\em inference phase} of PASE is executed when a
request for classification is received -- in this case, we select
the element of our constructed ensemble of classifiers that was {\em
not} trained on this input or the element of the training data that
is closest to that input and use it for classification of the
provided input (hence the term ``switching ensemble'').

\begin{figure}
  \centering
  \fbox{\includegraphics[width=0.45\textwidth]{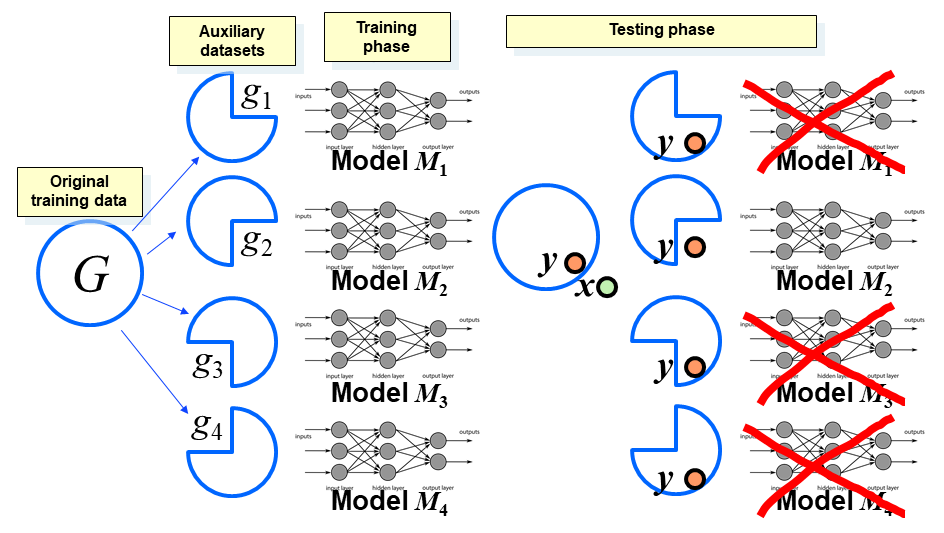}}
  \caption{PASE architecture.}
\label{PASE_ARCHITECTURE}
\end{figure}

More formally, Figure~\ref{PASE_ARCHITECTURE} shows that we generate
$k$ different auxiliary classifiers (for illustration purposes,
$k=4$ in Figure~\ref{PASE_ARCHITECTURE}), where each classifier is
built using an auxiliary subset of the original training data $G$
(shown as blue circle in Figure~\ref{PASE_ARCHITECTURE}), where
these subsets (shown as three-quarters of $G$ in
Figure~\ref{PASE_ARCHITECTURE}) are created in the same way as in
cross-validation procedure with $k=4$ folds. Specifically, we
randomly partition the set of training instances (elements of
training data $G$) into $k$ approximately equally-sized groups $g_1,
g_2,\ldots,g_k$. We then train $k$ classifiers $M_1,
M_2,\ldots,M_k$, where each $M_j$ is trained on all these groups
{\em except} group $g_j$, i.e., on $G\backslash g_j$. Our motivation
here is to create a set of these multiple classifiers so that, at
inference phase, we can select one of these classifiers that was
{\em not} trained on the given test instance $x$ (shown as a green
dot in Figure~\ref{PASE_ARCHITECTURE}). In order to ensure that,
before making the actual inference, we first select the element $y$
of the training data that is closest to test instance $x$ (for that,
we use Facebook AI Similarity Search (FAISS) open source
software~\cite{JDH17}), and then select the classifier that was
trained on a dataset that did not include $y$; in
Figure~\ref{PASE_ARCHITECTURE}, that is the classifier $M_2$.

\begin{figure}
  \centering
  \fbox{\includegraphics[width=0.45\textwidth]{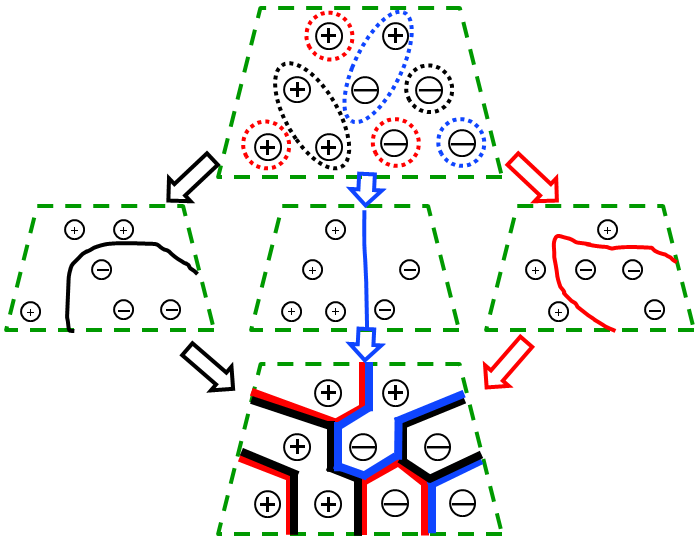}}
  \caption{PASE geometry.}
\label{PASE_GEOMETRY}
\end{figure}

In order to illustrate PASE operation geometrically, consider
Figure~\ref{PASE_GEOMETRY}. The upper third of
Figure~\ref{PASE_GEOMETRY} shows the training dataset consisting of 15
data points labeled as positive and negative, so the machine learning
task is to construct a binary classifier. Using $k=3$ in our PASE
approach, the training dataset is partitioned into three disjoint
auxiliary subsets shown in Figure~\ref{PASE_GEOMETRY} using black,
blue and red colors. The middle third of Figure~\ref{PASE_GEOMETRY}
shows $k=3$ auxiliary training subsets, each of which containing all
the original training data without one of the above
subsets. Specifically, the leftmost training dataset does not include
black subset, the middle training dataset does not include blue
dataset, and the rightmost training dataset does not include red
dataset. The auxiliary classifiers (black, blue, and red), trained on
each of the three auxiliary training datasets, form PASE switching
ensemble, which effectively partitions the space into multiple areas
(similar to Voronoi diagram) shown in the bottom part of
Figure~\ref{PASE_GEOMETRY}.  The switching mechanism of PASE then
selects the appropriate classifier for any given query point: for
instance, if the query point belongs to the area with black
boundaries, the black classifier will be applied; if the query point
belongs to the area with blue boundaries, the blue classifier will be
applied, etc.

As follows from the description above, if PASE is queried with a
data point $y$ from training data, a classifier that was {\em not}
trained on $y$ will be selected to respond to the query. Moreover,
as Figure~\ref{PASE_GEOMETRY} illustrates, this classifier will be
selected even the queried data point is not $y$ itself but rather a
point that is ``close'' to $y$ (i.e., located in the same area with
$y$). That property is especially important for image applications
since visual similarity to one of the elements of training data
might be actually quite sufficient for the purpose of membership
inference attack.

Due to the way PASE selects its classifiers for making an inference,
it might seem that a membership inference attack against PASE would
always label the data as being not in the training set, thus making
the true negative value of that binary attack classifier (the
membership attack system is essentially a binary classifier that
decides, based on the query response from the attacked system,
whether the given query was a part of training data or not) equal to
100\%. That is actually incorrect, and the confusion matrix of a
membership attack on a PASE-protected machine learning system
exhibits a more nuanced and interesting behavior. Conceptually, the
balance between false positives and false negatives of a membership
attack on PASE mostly depends on the sparsity of the training data
in the corresponding feature space.

Specifically, if data points are relatively ``close'' to each other
(or, to put it differently, the images do not exhibit much
individuality), the membership inference attack would tend to label
any point as belonging to the training set (which does not cause
privacy leakage since the training data points are poorly
distinguishable anyway). Conversely, if data points are relatively
``far'' from each other (or, to put it differently, the images do
exhibit significant individuality), the membership inference attack
would tend to label any point as {\em not} belonging to the training
set (which again does not cause privacy leakage for the training
data points). These effects match the observations made
in~\cite{Shokri} regarding the overall efficiency of membership
inference attacks: for instance, in CIFAR10 dataset~\cite{CIFAR10},
the privacy leakage for the class of cars is smaller than that of
the class of cats, while privacy leakage in MNIST
dataset~\cite{MNIST} is even smaller (images of cats apparently have
more ``individuality" that those of cars, and MNIST images of
heavily preprocessed handwritten digits have even less
``individuality''). Nevertheless, for the purpose of protection
against privacy leakage, this traditional tradeoff between false
positives and false negatives eventually yields the accuracy of the
membership inference attack (against PASE-protected system) being
practically equal to 50\% (i.e., the toss-up decision for the
attacker), as illustrated in the next Section on several
representative experiments.

In terms of practical implementation of the described PASE approach,
the same type / architecture of machine learning model that has been
already optimized for a given dataset can be used for training for
each of the ensemble classifiers -- there is no need to change that
architecture with the reasonable risk of adversely affecting the
overall performance. Depending on the number $k$ of algorithms in
the switching ensemble, the accuracy penalty of our PASE solution
can be made negligibly small (of course, the training time will have
to be proportionally increased), at least within the context of
membership inference attack of the type~\cite{Shokri}.

Since PASE inference process includes the selection of the
appropriate algorithm from the constructed switching ensemble, the
inference would take more time than the standard machine learning
model. However, in the experiments we have carried out so far
(described in the next Section), we have not observed any
practically meaningful delay in terms of PASE inference time.

To summarize, the proposed PASE approach
\begin{itemize}
\item prevents privacy leakage in the context of membership inference
attacks of the type~\cite{Shokri},
\item does not cause any meaningful deterioration of accuracy / utility of the
protected machine learning model,
\item does not require any architecture changes of the
machine learning model preferred for the given data
\end{itemize}
at the expense of practically acceptable
\begin{itemize}
\item increased training time,
\item increased inference time,
\item increased storage requirements for maintaining switching
ensemble models and training data index.
\end{itemize}
The next Section illustrates these tradeoff properties and provides
calibration results for a diverse set of image classification tasks.

Finally, note that PASE is designed for the type of current
state-of-the-art membership inference attack introduced
in~\cite{Shokri}. The efficiency of such practical attacks has been
demonstrated on a variety of diverse datasets, and their key enabler
(confidence overfitting on some elements of the training data) for
such privacy leakage is already well understood. Given PASE
protection against such attacks, it is theoretically possible that
an adversary might explore its architecture for launching the
following ``next level'' attack: by using significant amount of
probing the outputs of the machine learning model by feeding it
various data points in the feature space, the attacker can reverse
engineer the boundaries of the areas where PASE decision is switched
from one classifier to another (illustrated in the bottom third of
Figure~\ref{PASE_GEOMETRY}) and then infer the identity of the data
point(s) contained in these areas by some processing of these
boundaries.

However, the practicality of such an attack is yet unclear: besides
seemingly enormous amount of probing it would take to discover any
meaningful portion of boundaries (that amount can be probably
significantly reduced if the probing is carried out within a
reasonably low-dimensional manifold containing the training data --
for instance, as it has been done in the recently proposed model
inversion attack~\cite{Basu2019MembershipMI}, although current model
inversion attacks produce just generalized class representations,
which is very different from finer-granularity identities of
training data points), the way of identifying the training data
point from the discovered boundaries is unknown. The investigation
of practicality and efficiency of such kinds of model inference
attacks is thus a subject of future work. However, in the event that
these attacks turn out to be implementable and efficient, the
current PASE architecture can be then easily modified, for instance,
by using the ``moving target'' approach, where $k$ auxiliary subsets
are periodically changed (and the corresponding $k$ classifiers in
the PASE ensemble are retrained), thus completely invalidating the
results of the previous probes by the attacker. The exact details of
such modifications are also a subject of future work, to be carried
out depending on the feasibility of the ``next level'' attack
described above.

\section{Experiments and Discussion}

We have carried out several experiments for evaluating and comparing PASE and PATE~\cite{pate} approaches in terms of the membership
privacy attack~\cite{Shokri} designed to determine whether a given
data record was a member of the model's training dataset. In our
experiments, we used three datasets: \noindent {\bf CIFAR-10}
dataset~\cite{CIFAR10} contains 50,000 training images
and 10,000 test images.\newline \noindent{\bf MNIST}
dataset~\cite{MNIST} contains 60,000 training images and 10,000
test images.\newline \noindent {\bf SVHN} dataset~\cite{SVNH}
contains 600,000 images.

We re-partitioned these standard datasets in the following way: (1)
merged the training and test sets; (2) split the whole dataset into
two disjoint parts, where the first part is for the baseline model,
PASE model and PATE model, and the second part is for the membership
attack model; (3) the dataset for each model is further split to
training and test sets (for the training data of PATE model, we used
90\% for the teacher model's training and the remaining 10\% -- for
the student model's training).

We have used different classification models for different datasets
to show  the effects of the privacy preservation models.
For each dataset, we have used the same classification model for the
baseline (undefended) model, PASE model, PATE model and the shadow
models of the membership inference. For {\bf MNIST}, we used a 3-layer
fully connected neural network (DNN - Dense layer Neural Network).
For {\bf SVHN}, we
used a CNN model from the released PATE code which consists of 2
convolutional layers and some fully connected layers, pooling layers
and
normalization layers.  For {\bf CIFAR-10}, we used a VGG16 neural
network
architecture.  Finally, for the {\bf attack} model, we used a 3-layer
fully connected neural network
(DNN) similar to the one used for the MNIST dataset described above.

As~\cite{Shokri} demonstrated, even small overfitting can be
efficiently leveraged by an attacker towards making statistically
meaningful membership inference. Since generalization performance of
different machine learning algorithms vary, in order to demonstrate
both the efficiency of membership inference attack on the
unprotected machine learning model and the efficiency of
corresponding membership privacy protection mechanisms, we need to
choose a machine learning model that exhibits some overfitting while
having good utility accuracy. For example, on the model architecture
selection for the {\bf MNIST} dataset, we used the DNN model because
the DNN model has good utility accuracy and noticeable overfitting.
In one experiment of undefended baseline model training on {\bf
MNIST } dataset with DNN model, the test accuracy was 97.06\% and
the training accuracy was 100\%. The generalization error, i.e., the
difference between training and testing accuracy,
was thus 2.94\% which indicates noticeable overfitting. We also
tried the CNN model on {\bf MNIST } dataset. The testing accuracy
was very good (which was 99\%) but the overfitting was small (the
generalization error was 1\%). Therefore, we did not use the CNN
model for {\bf MNIST } dataset to explore the effects of the privacy
preservation models: there would not be much privacy leakage there,
and thus there would not be much need for prevention of such privacy
leakage.

Similarly, the generalization error of the baseline model with the
VGG16 architecture on {\bf CIFAR-10} dataset was 25.9\% and the
generalization error of the baseline model with the CNN architecture
on {\bf SVHN} dataset was 3.4\%. Note that the generalization errors
were higher than typical models trained with full datasets. We did
not use the full datasets for the training and testing of baseline
models as well as of PASE and PATE models because we put aside part
of the data for training the membership attack's shadow models.
Overfitting and generalization error usually increase with the
decrease of the size of training data.

During the inference phase, one of our $k$ auxiliary PASE
classifiers is selected to answer a test query. We assume that the
samples in the PASE training data are unique. If the querying sample
is in the training data, there will be only one classifier which is
not trained with that querying sample and that classifier is chosen
to answer that query. If the querying sample is not in the training
data, none of the  $k$ PASE classifiers are trained with that
querying sample. In that case, we look for a sample in the training
data which is most similar to the querying sample and choose the
classifier which was not trained with that most similar sample to
answer the query. Training data are randomly partitioned and similar
training samples may end up in random different partitions. If there
are samples in the training data that are not unique, we can either
remove the duplicate samples or partition the training data in such
a way that the same duplicate samples land in a same partition so
that PASE switching mechanism will still work properly. In our
experiments, we used the FAISS library ~\cite{JDH17} to build the
search index for similarity search of test sample on training data.
The index for fast searching can be precomputed and can be
efficiently used for very large datasets. For the experimental
results presented in this paper, we simply converted the training
images from their original 2D matrix format to 1D vectors and then
built the FAISS index using $L_2$ distance metric. Since image
similarity search can also rely on various image feature descriptors
/ representations, we plan to explore, in our future work,
alternative ways of building the search index by leveraging
specialized feature descriptors and representations of the images,
such as SIFT~\cite{siftDavid}, image embedding using Convolutional
Neural Networks ~\cite{kielaEmbeddings}, etc.

The results of our experiments are summarized in
Table~\ref{Results-utility} - Table~\ref{Inference-Results}. For
comparison purposes, the PATE results are produced without the
random noise added for the PATE model's student. Indeed, if it is
added, the privacy attack accuracy does not change (it is already at
the level of 50\%), while the utility accuracy is reduced. A large
number of teacher classifiers can then compensate the utility
accuracy loss, but, for some datasets such as {\bf CIFAR10 }, a
large number of teacher classifiers reduces the size of training
data for each PATE teacher, which, in turn, reduces
 the teacher model's accuracy, and, therefore, the student's
 accuracy will be also reduced.
 The PASE models used ensemble of
$k=5$ classifiers. The attack models used 10 shadow models for
training data generation. The other experiment parameters used for
the results in tables were as follows. \newline
\noindent {\bf CIFAR-10} dataset: 29,000 samples used for utility
model training and test; the utility models used the VGG16 neural
network; the PATE model used 10 teachers. \newline \noindent {\bf
MNIST} dataset: 30,000 samples used for utility model training and
test; the utility models used the DNN neural network; the PATE model
used 20 teachers.
\newline \noindent {\bf SVHN} dataset: 80,000 samples used for
utility model training and test; the
utility models used the CNN neural network; PATE model used 100
teachers.
\begin{table}[!htbp]
\begin{center}
\begin{tabular}{ | l | c | c | c |}
\hline &\multicolumn{3}{c|}{Utility accuracy}\\ \hline Dataset &
Baseline & PASE & PATE \\ \hline CIFAR10 & 74.13\% & 69.64\% &
26.79\% \\\hline MNIST & 97.06\% & 96.53\% & 88.87\% \\ \hline SVHN
& 95.46\% & 95.11\% & 79.73\% \\ \hline
\end{tabular}
\end{center}
\caption{Utility accuracy of baseline
classification model, PASE and PATE
privacy-preserving models.} \label{Results-utility}
\end{table}

\begin{table}[!htbp]
\begin{center}
\begin{tabular}{ | l | c | c | c |}
\hline &\multicolumn{3}{c|}{Attack accuracy}\\ \hline Dataset &
Baseline & PASE & PATE \\ \hline CIFAR10 & 68.70\% & 50.17\% &
50.20\% \\ \hline MNIST & 53.48\% & 50.15\% & 51.00\% \\ \hline SVHN
&  52.15\% & 49.87\% & 50.22\%\\ \hline
\end{tabular}
\end{center}
\caption{Membership inference attack accuracy of baseline
classification model, PASE and PATE
privacy-preserving models.} \label{Results-attack}
\end{table}

As Table~\ref{Results-attack} shows, both PATE and PASE approaches
achieve their goal of eliminating privacy leakage (the success rate
of membership inference attack is practically equal to the value
50\% of the random choice), with PASE approach maintaining the
accuracy of the defended model quite closely to the original
baseline model as shown in Table~\ref{Results-utility}. Note that it
was achieved by using only $k=5$ classifiers in PASE ensemble: a
proportionally longer training time of a larger number of $k$
classifiers would provide classification performance that is even
closer to the baseline model, while retaining the same level of
privacy protection.

\begin{table}[!htbp]
\begin{center}
\begin{tabular}{ | l | c | c | c | }
\hline &\multicolumn{3}{c|}{Training time  } \\
       &\multicolumn{3}{c|}{( the ratio over baseline training time) } \\
\hline Dataset & Baseline & PASE & PATE \\ \hline CIFAR10 & 1 & 3.3
&
1.2 (10 teachers) \\ \hline MNIST & 1 & 3.2 & 1.4 (20 teachers) \\
\hline SVHN & 1 & 3.1 & 23.2 (100 teachers) \\ \hline
\end{tabular}
\end{center}
\caption{Training time of baseline
classification model, PASE and PATE
privacy-preserving models.} \label{Timing-Results}
\end{table}

Table~\ref{Timing-Results} shows the computing time used to train
the  models.  The results  are the ratios of the models' training
time over their corresponding baseline models' training time. We
used $k=5$ in our experiments, and the total time to train the five
PASE models is about $3.2$ times larger than the time used to train
one baseline model. Note that the training time of PATE models
varies significantly since different numbers of teachers were used.

\begin{table}[!htbp]
\begin{center}
\begin{tabular}{ | l | c | c | c | c | } \hline
 & &\multicolumn{3}{c|}{Inference time } \\
 & &\multicolumn{3}{c|}{( millisecond per sample) } \\ \hline
Dataset & Architecture & Baseline & PASE & PATE \\ \hline CIFAR10 &
VGG16 & 0.35 & 0.94 & 0.30 \\ \hline MNIST & DNN & 0.034 & 0.120 &
0.028 \\ \hline SVHN & CNN & 0.107 & 0.416 & 0.106 \\ \hline
\end{tabular}
\end{center}
\caption{Inference time of baseline
classification model, PASE and PATE
privacy-preserving models.} \label{Inference-Results}
\end{table}

Table~\ref{Inference-Results} shows the inference time of the
trained models on their corresponding test samples.
 PASE model's inference time is expectedly longer
 (but still well within
 practical bounds) to that of the baseline model.
 This is mostly due to the time used to search for the  most
 similar training sample in the FAISS index. The PATE model's
 inference time is comparable to that of the baseline model.
 The VGG16 model, which was used for CIFAR10 dataset,
 is more complex than the CNN model which was used for
 the SVHN dataset; the CNN model is more complex
 than the DNN model, which was used for the MNIST dataset.
 As a result, the inference times of the CIFAR10 models
 were longer
 than those of the SVHN models, and the inference times of SVHN models
 were
 longer than those of the MNIST models.

\section{Conclusions}

We have proposed and tested a novel approach of preventing privacy
leakage of machine learning models during membership inference
attacks. We showed that our approach has good performance (both in
terms of privacy leakage avoidance and small accuracy penalty) on
several calibration image datasets.

In future work, we plan to explore various implementation options
(such as selection of number and type of constituent auxiliary
classifiers in the switching ensemble, choices in inference
mechanism, etc.), build similarity search index with state of the
art feature descriptors and explore the applicability of our
approach to non-image datasets. We also plan to investigate the
feasibility of ``next level'' model inversion attacks against PASE
architecture (as described in the end of Section 2) and the
corresponding modifications of PASE architecture that they might
necessitate.

\subsubsection*{Acknowledgments}

The authors are grateful for the support of NGA via contract number
NM0476-19-C-0007. The views, opinions and/or findings expressed are
those of the authors and should not be interpreted as representing
the official views or policies of the Department of Defense or the
U.S. Government. This paper has been approved for public release,
NGA \# 19-977.

\newpage
\medskip

\small

\bibliography{PASE_cvpr_2020}
\bibliographystyle{abbrv}

\end{document}